\documentclass[11pt,a4paper]{article}
\usepackage[hyperref]{acl2019}
\usepackage{times}
\usepackage{amsmath}
\usepackage{url}
\usepackage{color,soul}
\usepackage{todonotes}
\usepackage{multirow}
\usepackage{subcaption}
\usepackage{comment}
\usepackage[shortlabels]{enumitem}
\setlist[enumerate]{nosep}

\usepackage{url}
\aclfinalcopy

\newcommand{\Note}[2]{} 
\newcommand{\SideNote}[2]{} 
\renewcommand{\Note}[2]{\todo[color=#1,size=\small, inline=true]{#2}} 
\renewcommand{\SideNote}[2]{\todo[color=#1,size=\small]{#2}} %

\title{Contextualized Word Embeddings Enhanced Event Temporal Relation Extraction for Story Understanding}

\author{Rujun Han, Mengyue Liang, Bashar Alhafni, Nanyun Peng  \\
  \texttt{\{rujunhan, mengyuel, alhafni, npeng\}@usc.edu} \\
  University of Southern California \\
  Information Sciences Institute
  }

\begin{document}
\maketitle
\begin{abstract}
  Learning causal and temporal relationships between events is an important step towards deeper story and commonsense understanding. 
  Though there are abundant datasets annotated with event relations for story comprehension, many have no empirical results associated with them.  
  In this work, we establish strong baselines for event temporal relation extraction on two under-explored story narrative datasets: Richer Event Description (\textbf{RED}) and Causal and Temporal Relation Scheme (\textbf{CaTeRS}). To the best of our knowledge, these are the first results reported on these two datasets. We demonstrate that neural network-based models can outperform some strong traditional linguistic feature-based models. We also conduct comparative studies to show the contribution of adopting contextualized word embeddings (BERT) for event temporal relation extraction from stories. Detailed analyses are offered to better understand the results.
\end{abstract}

\section{Introduction}
Event temporal relation understanding is a major component of story/narrative comprehension. It is an important natural language understanding (NLU) task with broad applications to downstream tasks such as story understanding~\cite{charniak1972toward,winograd1972understanding,schubert2000episodic}, question answering~\cite{pustejovsky2003timeml,saquete2004splitting}, and text summarization ~\cite{allan2001temporal,georgescu2013temporal}.

\begin{figure}[t]
\centering
    \begin{subfigure}[b]{\columnwidth}
    \includegraphics[angle=-90,clip,trim=9cm 2.5cm 7cm 2.5cm, width=\columnwidth]{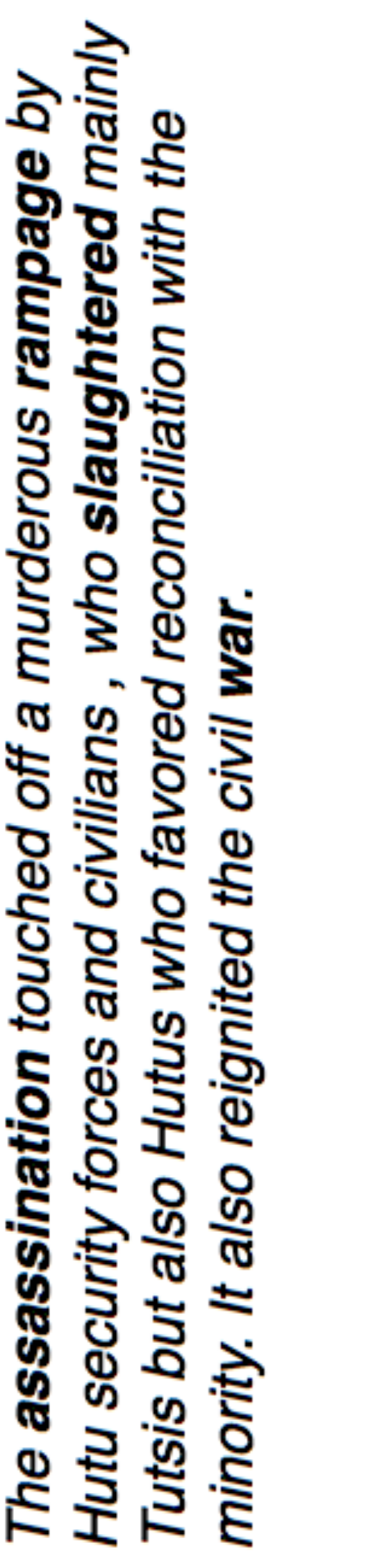}
    \vspace{-0.5cm}
    \end{subfigure}
    
    \begin{subfigure}[b]{0.8\columnwidth}
    \centering
    \includegraphics[clip,trim=0cm 7.5cm 0cm 7cm, width=0.8\columnwidth]{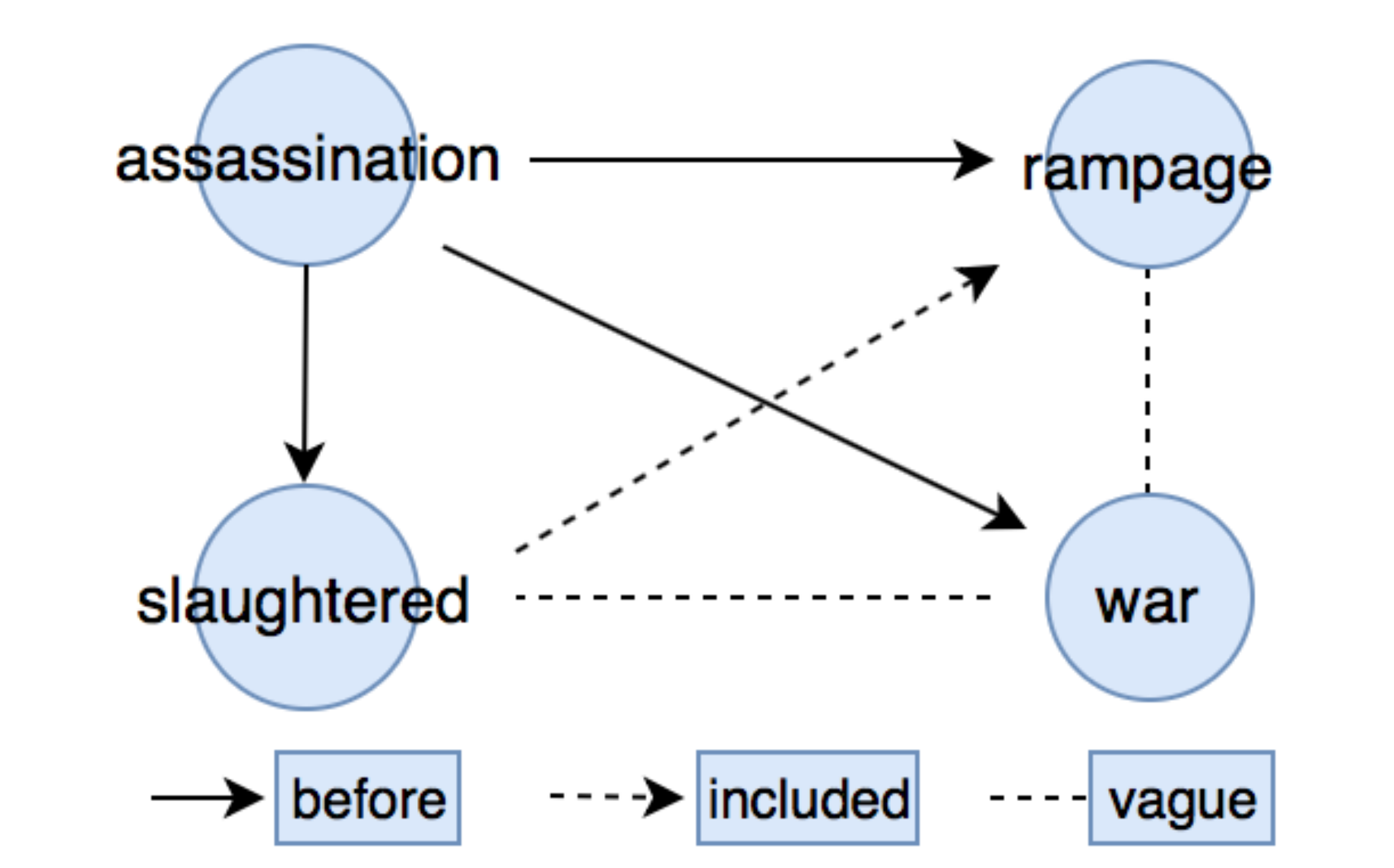}
    \label{fig:ex12}
    \end{subfigure}
    
\caption{An example paragraph with its (partial) temporal graphs. Some events are removed for clarity. 
}
\label{fig:ex1} 
\end{figure}

The goal of event temporal relation extraction is to build a directed graph where nodes correspond to events, and edges reflect temporal relations between the events.
Figure~\ref{fig:ex1} illustrates an example of such a graph for the text shown above. Different types of edges specify different temporal relations: the event \textbf{assassination} is before \textbf{slaughtered}, \textbf{slaughtered} is included in \textbf{rampage}, and the relation between \textbf{rampage} and \textbf{war} is vague. 

Modeling event temporal relations is crucial for story/narrative understanding and storytelling, because a story is typically composed of a sequence of events~\cite{MostafazadehX2016}. 
Several story corpora are thus annotated with various event-event relations to understand commonsense event knowledge. \textbf{CaTeRS}~\cite{MostafazadehGCAL2016} is created by annotating 320 five-sentence stories sampled from ROCStories~\citep{MostafazadehX2016} dataset. 
\textbf{RED}~\cite{O'Gorman2016} contains annotations of rich relations between event pairs for storyline understanding, including  co-reference and partial co-reference relations, temporal; causal, and sub-event relations. 

Despite multiple productive research threads on temporal and causal relation modeling among events~\cite{ChambersTBS2014, P18-1212, meng2018context} and event relation annotation for story understanding~\cite{MostafazadehGCAL2016}, the intersection of these two threads seems flimsy. 
To the best of our knowledge, no event relation extraction results have been reported on CaTeRS and RED. 

We apply neural network models that leverage recent advances in contextualized embeddings (BERT~\cite{devlin2018bert}) to event-event relation extraction tasks for CaTeRS and RED. Our goal in this paper is to increase understanding of how well the state-of-the-art event relation models work for story/narrative comprehension. 

In this paper, we report the first results of event temporal relation extraction on two under-explored story comprehension datasets: CaTeRS and RED. We establish strong baselines with neural network models enhanced by recent breakthrough of contextualized embeddings, BERT~\cite{devlin2018bert}. We summarize the contributions of the paper as follows:
\begin{enumerate}
    \item Establish strong neural network (NN) baselines for CaTeRS and RED datasets.  
    \item Comparison and analysis between NN and feature-based models.
    \item Demonstrate the effectiveness of leveraging contextualized word representation learning in event temporal relation extraction tasks.
\end{enumerate}

\begin{figure}[t]

\includegraphics[angle=-90,clip,trim=0cm 2.5cm 0cm 0cm,width=\columnwidth]{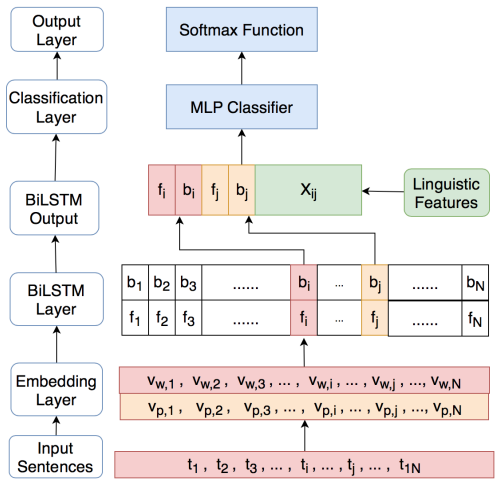}
\caption{
\label{fig:bisltm}
Deep neural network architecture for event relation prediction} 
\end{figure}  

\section{Models}
We investigate both neural network-based models and traditional feature-based models. We briefly introduce them in this section.
\paragraph{BiLSTM Classifier.} We adopt a recurrent neural network (RNN)-based pair-wise relation classifier in order to learn features in a data driven way and capture long-distance contexts in the input. The neural architecture is inspired by prior work in entity and event relation extraction such as ~\citet{tourille2017neural, cheng2017classifying, meng2017temporal, meng2018context}.

As shown in Figure~\ref{fig:bisltm}, the bottom layer corresponds to the input sentences.\footnote{Following the convention of event relation prediction literature \citep{ChambersTBS2014, P18-1212, NingWuRo18}, we only consider event pairs that occur in the same or neighboring sentences, but the architecture can be easily adapted to the case where inputs are longer than two sentences.}  We use indices $i$ and $j$ to denote the tokens associated with an event pair ($i, j$) $\in \mathcal{E}\mathcal{E}$ in the input sentences of length N.

Following the input layer, the embedding layer consists of both word and part-of-speech (POS) tag embeddings for each token, denoted as $v_{w,k}$ and $v_{p,k}$. We fix word embeddings while tuning the POS tag embeddings during model training. The word and POS embeddings are concatenated to represent an input token. They are then fed into a Bi-LSTM layer to get contextualized representations. We assume the events are annotated in the text. For each event pair ($i,j$), we take the forward and backward hidden vectors corresponding to each event, namely $f_i, b_i, f_j, b_j$ to encode the context of event tokens. 

Previous research demonstrated the success of leveraging linguistic features in event relation prediction. However, in an effort to reduce hand-crafted features, we only use token distance as a linguistic feature. In general, we denote linguistic features as $X_{i,j}$. Finally, all hidden vectors and linguistic features are concatenated to form the input to a final one-hidden-layer multi-layer perceptron (MLP) classifier to produce a softmax distribution over all possible pair-wise relations.
\paragraph{Feature-based Classifier.} We also compare our NN models with the current state-of-the-art feature-based model for event relation extraction CAEVO \citep{ChambersTBS2014}. CAEVO builds a pipeline with ordered sieves. Each sieve can either be a rule-based classifier or a machine learning model and sieves are sorted by model precision such that decisions from a lower precision classifier cannot contradict those from a higher precision model.
 
\section{Experimental Setup}
\subsection{Data}
\label{sec:data}
\paragraph{CaTeRS} is created by annotating 1600 sentences of 320 five-sentence stories sampled from ROCStories \citep{MostafazadehX2016} dataset. CaTeRS contains both temporal and causal relations in an effort to understand and predict commonsense relations between events. 

As demonstrated in Table~\ref{tab:data}, we split all stories into 220 training and 80 test.\footnote{Annotated CaTeRS data can be found here: \url{http://cs.rochester.edu/nlp/rocstories/CaTeRS/}. Note that we were only able to retrieve 300 stories of the 320 and hence our total number of pairs are slightly lower.} We do not construct the development set because the dataset is small. Note that some relations have compounded labels such as ``CAUSE\_BEFORE'', ``ENABLE\_BEFORE'', etc. We only take the temporal portion of the annotations. 

\paragraph{RED} 
annotates a wide range of relations of event pairs including their coreference and partial coreference relations, and temporal, causal and subevent relationships. 
We split data according to the standard train, development, test sets, and only focus on the temporal relations. 

\paragraph{Negative Pairs Construction.} The common issue of these two datasets is that they are not densely annotated -- not every pair of events is annotated with a relation. We provide one way to handle negative (unannotated) pairs in this paper. 
When constructing negative examples, we take all event pairs that occur within the same or neighboring sentences with no annotations, labeling them as ``NONE''. The negative to positive samples ratio is \textbf{1.00} and \textbf{11.5} for CaTeRS and RED respectively. Note that RED data has much higher negative ratio (as shown in Table~\ref{tab:data}) because it contains longer articles, more complicated sentence structures, and richer entity types than CaTeRS where all stories consist of 5 (mostly short) sentences. 

In both the development and test sets, we add all negative pairs as candidates for the relation prediction. During training, the number of negative pairs we add is based on a hyper-parameter that we tune to control the negative-to-positive sample ratio.

 \begin{table}[t]
 	\centering
 	\begin{tabular}{l|c|c} \hline
 	& CaTeRS & RED  \\ \hline\hline
    \multicolumn{3}{c}{\textbf{\# of Documents}} \\ \hline
 	Train & 220 & 76\\
 	Dev  &   N/A & 9\\
    Test &  80 & 10 \\\hline\hline
    \multicolumn{3}{c}{\textbf{\# of Pairs}} \\ \hline
 	Train & 1692 & 3336 \\
 	Dev & N/A   & 400 \\
 	Test & 743  & 473 \\
 	Negative & 2432  & 48427  \\\hline\hline
 	\end{tabular}
   	\caption{Data overview: the number of documents in CaTeRS refers to the number of stories. ``Negative'' denotes negative pairs (missing annotations) within two sentence span we construct for the whole dataset.}
   	\label{tab:data}
 \end{table}

\paragraph{Sentence Distance Distribution.}  To justify our decision of selecting negative pairs within the same or neighboring sentences, we show the distribution of distances across positive sentence  pairs in Table~\ref{tab:sent_dist}. Although CaTeRS data has pair distance more evenly distributed than RED, we observe that the vast majority (85.87\% and 93.99\% respectively) of positive pairs have sentence distance less than or equal to one. 

To handle negative pairs that are more than two sentences away, we automatically predict all out-of-window pairs as ``NONE''. This means that some positive pairs will be automatically labeled as negative pairs. Since the percentage of out-of-window positive pairs is  small, we believe the impact on performance is small. We can investigate expanding the prediction window in future research, but the trade-off is that we will get more negative pairs that are hard to predict.

  \begin{table}[t]
 	\centering
 	\begin{tabular}{l|c|c} \hline
 	& CaTeRS & RED  \\ \hline\hline
    \multicolumn{3}{c}{\textbf{Token Sentence Distance (\%) }} \\ \hline
 	0 & 39.20 & 88.24\\
 	1  & 46.67 & 5.75\\
 	2 & 7.02 & 2.00\\
 	3  & 3.53 & 0.83\\
    $\ge$ 4 & 3.57 & 3.41 \\\hline
 	\end{tabular}
   	\caption{Token sentence distance breakdown. \textbf{0}: a pair of events in the same sentence; \textbf{1}: a pair of events in the neighboring sentence (2 sentence span); \textbf{2}: a pair of events in 3 sentence span, etc.}
   	\label{tab:sent_dist}
 \end{table}
 
\subsection{Implementation Details}
\paragraph{CAEVO Baseline.} CAEVO consists of both linguistic-rule-based sieves and feature-based trainable sieves. We train CAEVO sieves with our train set and evaluate them on both dev and test sets.\footnote{CAEVO implementation can be found here: https://github.com/nchambers/caevo.} CAEVO is an end-to-end system that automatically annotates both events and relations. In order to resolve label annotation mismatch between CAEVO and our gold data, we create our own final input files to CAEVO system. Default parameter settings are used when running the CAEVO system.

\paragraph{NN Classifier with GloVe Embedding.} In an effort of building a general model and reducing the number of hand-crafted features, we leverage pre-trained (GloVe 300) embeddings in place of linguistic features. The only linguistic feature we use in our experiment is token distance. We notice in our experiments that hidden layer size, dropout ratio and negative sample ratio impact model performance significantly. We conduct grid search to find the best hyper-parameter combination according to the performance of the development set. 

Note that since the CaTeRS data is small and there is no standard train, development, and test splits, we conduct cross-validation on training data to choose the best hyper-parameters and predict on test. For RED data, the standard train, development, test splits are used.

\paragraph{NN Classifier with BERT Embedding.} As we mentioned briefly in the introduction, using BERT output as word embeddings could provide an additional performance boost in our NN architecture. We pre-process our raw data by feeding original sentences into a pre-trained BERT model\footnote{We use pre-trained BERT-Base model with 768 hidden size, 12 layers, 12 heads implemented by \url{https://github.com/huggingface/pytorch-pretrained-BERT}} and output the last layer of BERT as token representations. In this experiment, we fix the negative sample ratio according to the result obtained from the previous step and only search for the best hidden layer size and dropout ratio.

 \begin{table}[t]
 	\centering
 	\begin{tabular}{l|c|c} \hline
 	& CaTeRS & RED  \\ \hline\hline
    \multicolumn{3}{c}{\textbf{NN Model - GloVe Embedding}} \\ \hline
 	hid\_size & 40 &  50\\
 	dropout & 0.6 & 0.3\\
 	neg\_ratio & 0.5 & 8.0\\
 	lr & 0.0005 &  0.0005\\\hline\hline
\multicolumn{3}{c}{\textbf{NN Model - BERT Embedding}} \\ \hline 
 	hid\_size & 50 & 100\\
 	dropout & 0.5 & 0.4\\
 	lr & 0.0005 & 0.0005\\\hline
 	\end{tabular}
   	\caption{Best hyper-parameters: \textbf{C}: controls for the strength of L1 penalty; \textbf{balanced}: is a binary indicator of whether training on ``balanced'' labels; \textbf{max\_iter}: early stopping criteria.}
   	\label{tab:hyper}
 \end{table}

 \begin{table}[t]
 	\centering
 	\begin{tabular}{l|c|c} \hline
 	& CaTeRS & RED \\ \hline\hline
    \multicolumn{3}{c}{\textbf{CAEVO}} \\ \hline
    Dev & NA  &  0.270 \\
 	Test & 0.396 &   0.274 \\\hline\hline
    \multicolumn{3}{c}{\textbf{NN Model - GloVe Embedding}} \\ \hline
 	Dev & 0.506 &  0.311 \\
 	Test & 0.476 &  0.308\\ \hline\hline
 	\multicolumn{3}{c}{\textbf{NN Model - BERT Embedding}} \\ \hline
 	Dev &  \textbf{0.527} &  \textbf{0.342} \\
 	Test &  \textbf{0.519} & \textbf{0.340}  \\ \hline
 	\end{tabular}
   	\caption{F1 Scores on development and test set for the two datasets. Note for CaTeRS data, we didn't conduct cross-validation on CAEVO, but instead train the model with default parameter settings. Hence the dev performance doesn't apply here.}
   	\label{tab:results}
 \end{table}

\begin{figure}[t]
\includegraphics[width=\columnwidth]{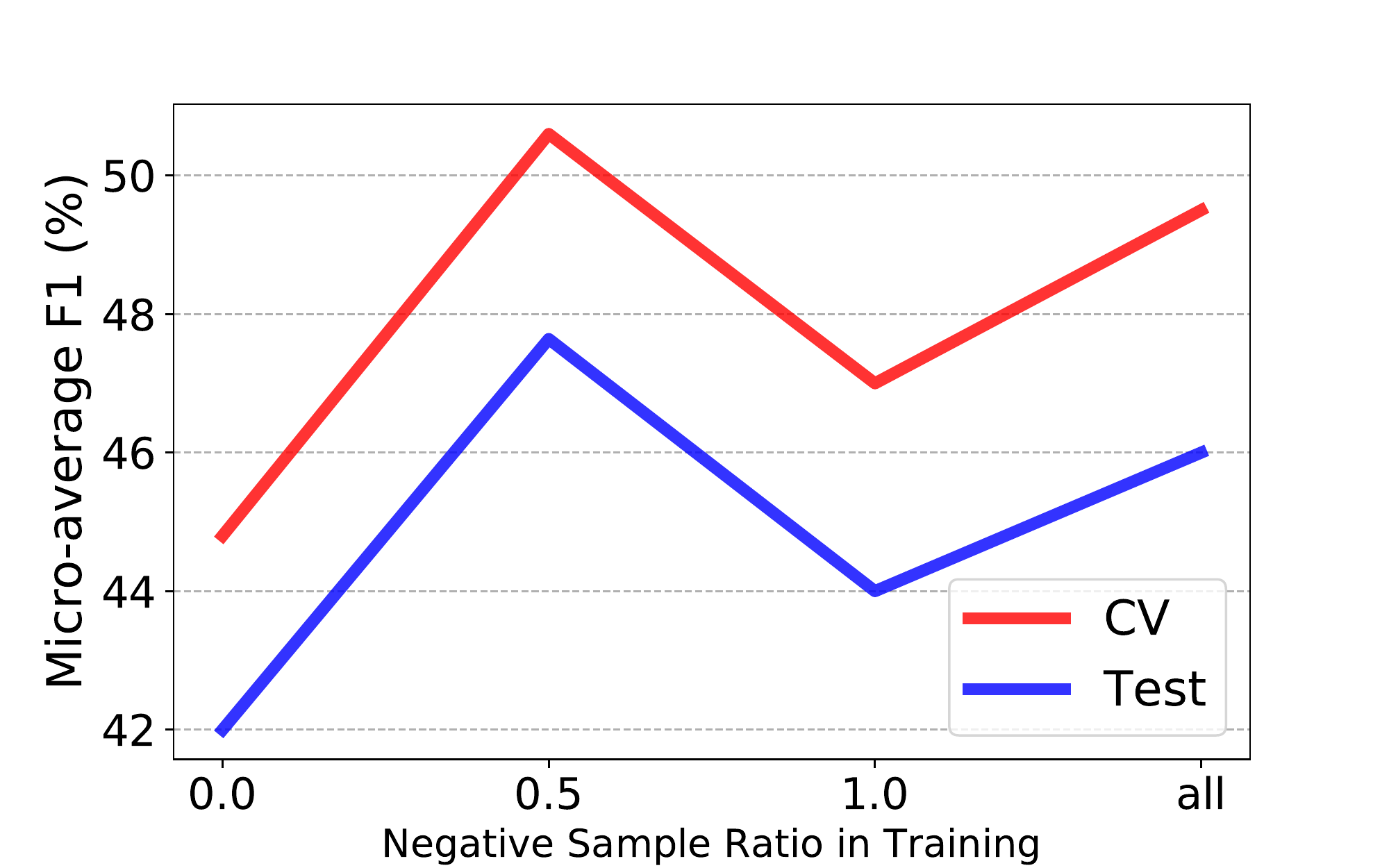}
\caption{
\label{fig:cat_cmp}
NN model (with GloVe embedding) performance with different negative sample ratio for CaTeRS.} 
\end{figure}

\begin{figure}[t]
\includegraphics[width=\columnwidth]{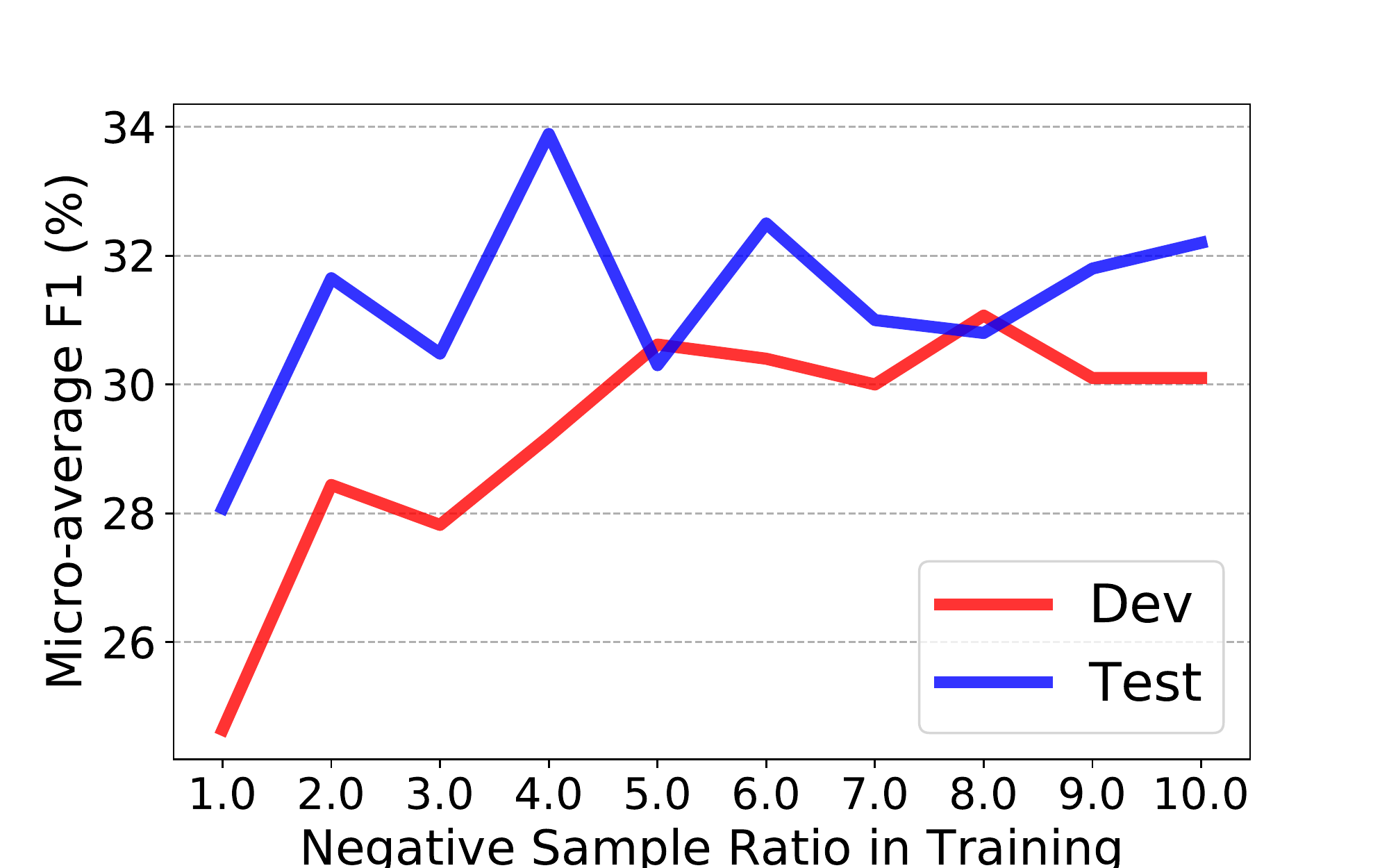}
\caption{NN model (with GloVe embedding) performance with different negative sample ratio for RED.}\label{fig:red_cmp}
\end{figure}

\section{Result and Analysis} Table~\ref{tab:hyper} contains the best hyper-parameters and Table~\ref{tab:results} contains micro-average F1 scores for both datasets on dev and test sets. We only consider positive pairs, i.e. correct predictions on NONE pairs are excluded for evaluation. In general, the baseline model CAEVO is outperformed by both NN models, and NN model with BERT embedding achieves the greatest performance. We now provide more detailed analysis and discussion for each dataset.

\paragraph{Performances on CaTeRS.} As we can observe in Table~\ref{tab:results}, with GloVe embeddings, the NN model outperforms CAEVO model by a wide margin. Note that our goal is not to show that our neural network model outperforms all feature-based models; instead, we aim at establishing strong event relation extraction baselines for story comprehension datasets.

In Figure~\ref{fig:cat_cmp}, we show that GloVe NN model achieves best cross validation (CV) and test F1 scores when negative to positive sample ratio is set to 0.5 in the training process. We observe that both CV and test performance degrades when using more negative samples.

With BERT embeddings, we observe 2.1\% and 4.3\% absolute performance improvement over development and test sets respectively. This demonstrates the effectiveness of contextualized representation learning by replacing GloVe embeddings with BERT representations.

\paragraph{Performances on RED.} In Table~\ref{tab:results}, We observe similar pattern for the RED dataset. The NN model with GloVe embeddings outperforms CAEVO by more than 3\% and the NN model with BERT embeddings further increases F1 scores by 3.1\% and 3.2\% per absolute scale for the dev and test sets respectively. 

Figure~\ref{fig:red_cmp} shows that the best F1 score for the dev set is achieved by setting the negative-to-positive sample ratio to 8.0 in training the NN-GloVe model. The best test F1 corresponding to this ratio is 30.8\%. 

Note that the performances on RED are much worse than those on CaTeRS. As we mentioned, we create negative pairs within a 2-sentence span. However, RED data has much larger size of negative pairs as shown in Table~\ref{tab:data}. Negative pairs are extremely hard to predict as they could either be no-relation pairs or their true relations may have been ignored by annotators. The latter introduces noise during model training as positive pairs are incorrectly labeled as ``NONE''. We suspect this is the main contributing factor to the large performance gap between CaTeRS and RED data.

\paragraph{Error Analysis.} For error analysis, we leverage CaTeRS data to show how BERT embeddings outperforms GloVe in our task. 

\begin{table}[t]
    \centering
    \begin{tabular}{c|c|c}
        \hline
         & \textbf{GloVe} & \textbf{BERT}  \\
         \hline
         \textbf{Precision} & 0.453 & 0.456 \\ \textbf{Recall} & 
         0.502 & 0.602 \\
         \textbf{F1} &
         0.476 & 0.519 \\
         \textbf{\# Correct} & 373 & 447 \\
         \hline
    \end{tabular}
    \caption{NN performances with GloVe and BERT embeddings respectively.}
    \label{tab:glove_bert}
\end{table}

Table~\ref{tab:glove_bert} shows results on test sets of the two datasets with more details. As we can observe, the improvement of F1 score is mainly contributed by 10\% increase of recall score. 
In Table~\ref{tab:examples}, we pick several examples amongst this category and qualitatively analyze them in the following paragraph.

\begin{table*}[t]
    \centering
    \begin{tabular}{|p{95mm}|c|c|}
     \hline
     \multicolumn{1}{|c|}{\bfseries Sentence} & {\bfseries True Relation} & {\bfseries GloVe Prediction} \\ \hline\hline
     1. Nick \textbf{was from} a poor family.
     \newline He wanted to attend college, but they just couldn't \textbf{afford} it. & \textit{OVERLAP} & \textit{NONE} \\
     \hline
     2. Rosie \textbf{wakes} up with very dry and painful eyes.
     \newline Even after removing her contacts, her eyes are still \textbf{irritated}. & \textit{OVERLAP} & \textit{NONE} \\\hline
     3. Lucy \textbf{awoke} and jumped out of bed.
     \newline It \textbf{was Independence Day} and she was excited. & \textit{OVERLAP} & \textit{BEFORE} \\
     \hline
     4. For six solid months she \textbf{walked} miles \textbf{delivering} papers. & \textit{OVERLAP} & \textit{BEFORE} \\\hline
     5. The customer kept \textbf{demanding} a drink that didn't exist. \newline Eventually, Cathy just \textbf{gave} her a latte. & \textit{BEFORE} & \textit{NONE} \\
     \hline
\end{tabular}
\caption{Examples of temporal relations misclassified with GloVe embedding but correct with BERT embedding.}
\label{tab:examples}
\end{table*}

First, BERT embdding can have better performance than GloVe when two events are far away from each other. Event pairs in sentences 1 and 2 are examples falling into such category. This shows that BERT is better at capturing long-term dependency. Moreover, GloVe embedding is pre-trained on text corpus based on co-occurrence statistics, while BERT incorporating contextual information during both training and inference. Event pairs in sentences 3, 4 and 5 demonstrate the importance of understanding of contextual information in temporal relation reasoning. For example, in sentence 3, the phrase ``Independence Day'' is an event that lasts for a period, whereas the action ``awoke'' is an instant action. BERT embedding recognizes that ``awoke'' occurs in the context of ``Independence Day'' and hence successfully predicts the relation between this pair of events as OVERLAP.

\section{Related Work} 
\subsection{Temporal  Relation  Data}
Collecting dense TempRel corpora with event pairs fully annotated has been reported challenging since annotators could easily overlook some pairs \citep{P14-2082, Bethard:2007:TTI:1304608.1306306, ChambersTBS2014}. TimeBank \citep{PustejovskyX2003} is an example with events and their relations annotated sparsely.  \textbf{TB-Dense} dataset mitigates this issue by forcing annotators to examine all pairs of events within the same or neighboring sentences. However, densely annotated datasets are relatively small both in terms of number of documents and event pairs, which restricts the complexity of machine learning models used in previous research. 

\subsection{Feature-based Models}
The series of TempEval competitions \citep{Verhagen:2007:STT:1621474.1621488, Verhagen:2010:STT:1859664.1859674, S13-2001} have attracted many research interests in predicting event temporal relations. 
Early attempts by \citet{Mani:2006:MLT:1220175.1220270, Verhagen:2007:STT:1621474.1621488, Chambers:2007:CTR:1557769.1557820,Verhagen:2008:TPT:1599288.1599300} only use pair-wise classification models. State-of-the-art local methods, such as ClearTK \citep{S13-2002}, UTTime \citep{laokulrat-EtAl:2013:SemEval-2013}, and NavyTime \cite{chambers:2013:SemEval-2013} improve on earlier work by feature engineering with linguistic and syntactic rules. As we mention in the Section 2, CAEVO is the current state-of-the-art system for feature-based temporal event relation extraction \citep{ChambersTBS2014}. It's widely used as the baseline for evaluating TB-Dense data. We adopt it as our baseline for evaluating CaTeRS and RED datasets. Additionally, several models \newcite{BramsenDLB2006, ChambersJ2008, DoLuRo12, NingWuRo18, P18-1212} have successfully incorporated global inference to impose global prediction consistency such as temporal transitivity.

\subsection{Neural Network Model}

\paragraph{BiLSTM Classifier.} Neural network-based methods have been employed for event temporal relation extraction~\cite{tourille2017neural, cheng2017classifying, meng2017temporal, meng2018context} which achieved impressive results. However, the dataset they focus on is \textbf{TB-Dense}. We have explored neural network models on CaTeRS and RED, which are more related to story narrative understanding and generation. 

\paragraph{BERT Features.} In our NN model, we also leverage Bidrectional Encoder Representations from Transformers (BERT) \citep{BERT2018} which has shown significant improvement in many NLP tasks by allowing fine-tuning of pre-trained language representations. Unlike the Generative Pre-trained Transformer (OpenAI GPT) \citep{GPT2018}, BERT uses a biderctional Transformer \citep{Transformer2018} instead of a unidirectional (left-to-right) Transformer to incorporate context from both directions. As mentioned earlier, we do not fine-tune BERT in our experiments and simply leverage the last layer as our contextualized word representations.

\section{Conclusion}
We established strong baselines for two story narrative understanding datasets: CaTeRS and RED. We have shown that neural network-based models can outperform feature-based models with wide margins, and we conducted an ablation study to show that contextualized representation learning can boost performance of NN models. Further research can focus on more systematic study or build stronger NN models over the same datasets used in this work. Exploring possibilities to directly apply temporal relation extraction to enhance performance of story generation systems is another promising research direction.

\section*{Acknowledgement}
We thank the anonymous reviewers for their constructive comments, as well as the members of the USC PLUS lab for their early feedback.
This work is supported by Contract W911NF-15-1-0543 with the US Defense Advanced Research Projects
Agency (DARPA).


\bibliography{acl2019}
\bibliographystyle{acl_natbib}
\end{document}